\documentclass{article}

\usepackage{microtype}
\usepackage{graphicx}
\usepackage{subfigure}
\usepackage{booktabs}

\usepackage{tikz} 
\usepackage{enumitem}
\usepackage{csquotes}
\usepackage{placeins}

\usepackage{xspace}

\newcommand{\ie}{i.\,e.\xspace}
\newcommand{\eg}{e.\,g.\xspace}
\usepackage{amsmath,amsthm,amsfonts,dsfont}
\DeclareMathOperator{\Ff}{\mathcal{F}}
\DeclareMathOperator{\Xx}{\mathcal{X}}
\DeclareMathOperator{\R}{\mathbb{R}}
\DeclareMathOperator{\E}{\mathbb{E}}
\newcommand*{\indicator}[1]{\mathds{1}_{#1}}
\newcommand*{\set}[1]{\{#1\}}
\DeclareMathOperator{\Vv}{\mathcal{V}}
\DeclareMathOperator{\Hh}{\mathcal{H}}

\newtheorem{theorem}{Theorem}[section]
\theoremstyle{definition}
\newtheorem{definition}[theorem]{Definition} 
\newtheorem{lemma}[theorem]{Lemma}

\theoremstyle{definition}

\numberwithin{equation}{section} 

\usepackage{hyperref}

\usepackage{cleveref}

\usepackage[accepted]{icml2019}

\icmltitlerunning{Sample Complexity Bounds for RNNs with Application to Combinatorial Graph Problems}

\begin{document}

\twocolumn[
\icmltitle{Sample Complexity Bounds for Recurrent Neural Networks\\ with Application to Combinatorial Graph Problems}

\icmlsetsymbol{equal}{*}

\begin{icmlauthorlist}
\icmlauthor{Nil-Jana Akpinar}{cmu}
\icmlauthor{Bernhard Kratzwald}{eth}
\icmlauthor{Stefan Feuerriegel}{eth}
\end{icmlauthorlist}

\icmlaffiliation{cmu}{Department of Statistics and Data Science, Carnegie Mellon University, Pittsburgh, USA}
\icmlaffiliation{eth}{Chair of Management Information Systems, ETH Zurich, Switzerland}

\icmlcorrespondingauthor{Nil-Jana Akpinar}{nakpinar@andrew.cmu.edu}

\icmlkeywords{Machine Learning, ICML}

\vskip 0.3in
]

\printAffiliationsAndNotice{}

\begin{abstract}
Learning to predict solutions to real-valued combinatorial graph problems promises efficient approximations. As demonstrated based on the NP-hard edge clique cover number, recurrent neural networks (RNNs) are particularly suited for this task and can even outperform state-of-the-art heuristics. However, the theoretical framework for estimating real-valued RNNs is understood only poorly. As our primary contribution, this is the first work that upper bounds the sample complexity for learning real-valued RNNs. While such derivations have been made earlier for feed-forward and convolutional neural networks, our work presents the first such attempt for recurrent neural networks. Given a single-layer RNN with $a$ rectified linear units and input of length $b$, we show that a population prediction error of $\varepsilon$ can be realized with at most $\tilde{\mathcal{O}}(a^4b/\varepsilon^2)$ samples.\footnote{The $\tilde{\mathcal{O}}(\cdot)$ notation indicates that poly-logarithmic factors are ignored.} We further derive comparable results for multi-layer RNNs. Accordingly, a size-adaptive RNN fed with graphs of at most $n$ vertices can be learned in $\tilde{\mathcal{O}}(n^6/\varepsilon^2)$, i.\,e., with only a polynomial number of samples. For combinatorial graph problems, this provides a theoretical foundation that renders RNNs competitive.
\end{abstract}

\section{Introduction}
Generalization bounds for neural networks present an active field of research dedicated to understanding what drives their empirical success from a theoretical perspective. A widespread approach to statistical generalization is based on the sample complexity of learning algorithms, which refers to the minimum number of samples required in order to learn a close-to-optimal model configuration. Because of that, considerable effort has been spent on deriving sample complexity bounds for various neural networks. Recent examples of bounding sample complexities involve, for instance, binary feed-forward neural networks \citep{harvey_nearly-tight_2017} and convolutional neural networks \citep{du_how_2018}. However, similar results are lacking for real-valued recurrent neural networks~(RNNs). This is surprising, since RNNs find widespread adoption in practice such as in natural language processing \citep[\eg,][]{socher2011parsing}, when other unstructured data serves as input \citep[\eg,][]{graves2009novel}, or even when solutions to combinatorial graph problems are predicted \citep[\eg,][]{wang1996rnn_tsp}. A particular benefit is the flexibility of RNNs over feed-forward neural network \citep[\eg,][]{rumelhart_rnn_1986}: even when processing input of variable size, their parameter space remains constant. Hence, this dearth of theoretical findings motivates our research questions: \emph{What is the sample complexity of learning recurrent neural networks? How can upper bounding the sample complexity of RNNs be applied in practical use cases?}

\subsection{Contributions}

\textbf{Generalization abilities of recurrent neural networks.}  Our theoretical findings on RNNs allow us to make contributions as follows:

\begin{enumerate}[topsep=0pt]
	\item[(1)] \emph{Sample complexity bounds for RNNs.} We derive explicit sample complexity bounds for real-valued single and multi-layer recurrent neural networks. Specifically, we find that the estimation of a RNN with a single hidden layer, $a$ recurrent units, inputs of length at most $b$, and a single real-valued output unit requires only $\tilde{\mathcal{O}}(a^4b/\varepsilon^2)$ samples in order to attain a population prediction error of $\varepsilon$. In comparison, a $d$-layer architecture requires at most $\tilde{\mathcal{O}}(d^2a_{\max}^4b/\varepsilon^2)$ samples where $a_{\max}$ is the width of the first (or widest) layer. During our derivations, we focus on rectified linear units (ReLUs) as these are widespread in machine learning practice.\\
	To the best of our knowledge, our work is the first attempt to establish generalization and sample complexity bounds for real-valued recurrent neural networks. 
\end{enumerate}
\newpage
As a cautionary note, sample complexity (as defined here) presents a rate of convergence not towards obtaining the true labels for a combinatorial problem, but towards a statistically consistent estimation of the network parameters. Accordingly, the sample complexity of a neural network model presents a lower bound on the number of samples required in order to choose the best functional approximation that a given network can attain.

\textbf{Neural networks for combinatorial graph problems.} We demonstrate the value of our sample complexity bounds in an intriguing application: we study size-adaptive RNNs for predicting solutions to real-valued combinatorial graph problems. This constitutes a model-agnostic and yet powerful approach in order to to approximate solutions to combinatorial problems. 

The use of neural networks has been suggested as heuristic for combinatorial graph problems \citep{aiyer_theoretical_1990,hopfield_neural_1985}.\footnote{See our supplementary materials for a detailed overview on the topic. We also refer to \citet{smith_neural_1999,syed_neural_2013} for extensive overviews on the use of neural network models for combinatorial problems.} These network learn approximations to the actual task from labeled data, thereby achieving a generalizable approximation to the true solution. The resulting solver promises an efficient computation scheme for combinatorial problems with clear benefits: it requires only a fixed evaluation cost when being applied to unseen data, and obtains solution schemes, even when any form of heuristic is unknown. As we shall see later, recurrent neural networks can be highly effective at such tasks \citep[cf. also][]{wang1996rnn_tsp}. However, this approach lacks theoretical guarantees, foremost number of samples required to attain generalization properties. Upper bounding the sample complexity would provide a mathematical stopping criterion to the size of the training set and, in a broader view, could provide insights into the applicability of neural networks for combinatorial graph problems:

\begin{enumerate}
	\itemsep0em 
	\item[(2)] \emph{Theoretical findings.} Denoting the maximum number of nodes in input graphs $n$, the sample complexity of our RNN is bounded by $\tilde{\mathcal{O}}(n^6/\varepsilon^2)$. Accordingly, it reveals that only polynomial number of samples is required to choose a close-to-optimal network configuration. This renders our novel approach competitive to other problem-specific heuristics for combinatorial graph problems. As an impact to machine learning practice, it might be interesting to know that sample complexity grows polynomial with $n$, even when the solution space grows especially (as, \eg, for NP-hard problems).
	\item[(3)] \emph{Numerical validation.}	The theoretical findings are complemented by numerical experiments that analyze the approximation performance. Here we confirm that state-of-the-art neural networks can yield a satisfactory performance with a considerably smaller number of training samples. Here we draw upon the edge clique cover number problem, as this presents an NP-hard combinatorial graph problem for which even powerful heuristics are scarce. We find that the RNN approach can outperform the numerical performance of the only known heuristic from the literature \citep{kellerman_determination_1973}. Hence, this also motivates the theoretical study of RNNs as done in this work.
\end{enumerate}

\subsection{Comparison with existing work on generalization bounds for neural networks}

\textbf{Network size as a complexity measure.} Theoretical research into generalization ability and sample complexity of neural networks has received considerable traction in recent years \citep[\eg,][]{du_how_2018,neyshabur_exploring_2017,bartlett_spectrally-normalized_2017}. A large share of the various approaches relies upon some kind of complexity measure for the defined functional space or the underlying set of parameters. For instance, one line of work draws upon different norms of parameters for capacity measurement \citep{bartlett_sample_1998, bartlett_spectrally-normalized_2017, neyshabur_search_2014}. A different approach is to measure the so-called sharpness \citep{keskar_large-batch_2016}; \ie, the robustness of errors to changes in parameters. In contrast to these ideas, we rely upon a more straightforward understanding of complexity given by traditional shattering dimensions and network size, \ie, the number of parameters \citep{anthony_neural_2009}. This choice has several benefits: (1)~It yields explicit sample complexity bounds. (2)~It provides a uniform notion of complexity which enables us to make very general, distribution-independent statements. (3)~The combinatorial structure enables us to derive the first known bounds for recurrent architectures.

\textbf{Binary vs. real-valued output.} Prior research on network-size-dependent generalization ability focused primarily on the binary case covered by the analysis of the VC-dimension \citep[\eg,][]{bartlett_vapnik-chervonenkis_2003,harvey_nearly-tight_2017}. The previous stream of literature concentrates almost exclusively on feed-forward architectures, with a few notable exceptions \citep{dasgupta_sample_1996,koiran_vapnik-chervonenkis_1998} that are concerned with binary output. \citet{dasgupta_sample_1996} derive bounds on the VC-dimension of recurrent perceptron mappings, whereas \citet{koiran_vapnik-chervonenkis_1998} study recurrent architectures with sigmoid, polynomial, and piecewise polynomial activation functions. Both works are limited to fairly trivial architectures with single recurrent layers and which further neglect output units by outputting the sign of the first entry of their hidden state. Only very few contributions have been made to the real-valued case \citep{anthony_neural_2009}, and these works concentrate entirely on feed-forward networks but not on RNNs.
\\ 
\\
\textbf{Recurrent/non-recurrent architectures.} Bounds to the generalization error have been developed for different non-recurrent network architectures. \citet{arora_stronger_2018} suggest a compression framework for re-parametrization of trained networks and, thereby, yield new generalization bounds for deep feed-forward and convolutional networks. \citet{du_how_2018} derive sample complexity bounds for the learning of convolutional neural networks. Yet, to the best of our knowledge, no prior effort was put into generalization and sample complexity bounds of real-valued RNNs.

\textbf{Deterministic vs. probabilistic bounds.} Besides traditional PAC analysis, an alternative approach for deriving generalization bounds of neural networks is PAC-Bayes analysis \citep{mcallester_pac-bayesian_1998,langford_not_2002}. PAC-Bayes analysis employs random noise in order to determine the sensitivity of single parameters, which has been used in order to derive probabilistic generalization bounds for feed-forward neural networks \citep{neyshabur_pac-bayesian_2017,dziugaite_computing_2017} and convolutional neural networks \citep{pitas_pac-bayesian_2017}. However, the resulting sample complexity bounds are often not on par with traditional approaches from using the network size \citep{arora_stronger_2018}. Moreover, the additional source of randomness impedes statements, as they are probabilistic and thus non-deterministic. In contrast, this paper advocates a traditional PAC analysis based on the network size, which provides us with the benefit of deterministic statements. 

\subsection{Organization}

This paper is organized as follows. In \Cref{sec:theory}, we specify the problem of learning real-valued RNN function classes and, on top of that, derive novel bounds for its sample complexity. In \Cref{sec:eccn}, we demonstrate how the theoretical results can be leveraged in applications: we prove that neural learning of single-valued combinatorial graph problems can be achieved with a polynomial rate of consistency. \Cref{sec:experiments} provides additional numerical experiments, which reveal a favorable approximation capacity and further point out strengths of RNNs over feed-forward variants. \Cref{sec:conclusions} concludes with a summary. A detailed overview on neural optimization and detailed derivations for all proofs are relegated to the supplementary materials.

\section{Learning of real-valued RNN}
\label{sec:theory}

Notably, the below sample complexity bounds are independent of the problem distribution $P$ and, hence, all findings generalize to a wide range of learning tasks. In this sense, our subsequent analysis provides a worst-case analysis for general learning tasks and, in a later section, we demonstrate the benefits by showing an application to neural learning of combinatorial graph problems.

\subsection{Mathematical preliminaries}

\textbf{Statistical setting.} In general, a recurrent neural network with a single real-valued output unit (after appropriate rescaling) presents a parametrized class of functions, $\Ff=\{f_\theta:\Xx\to[0,1]:\theta\in\Theta\}$, where $\Xx$ denotes the set of real-valued input vectors. Given a distribution of labeled samples $(X_1,Y_1),\ldots,(X_n,Y_n)\sim P$, there is a sense of global optimality of the network configuration which is attained by minimizing the \emph{population error}. For the mean squared error, this is given by $\theta^\ast=\arg\inf_{\Theta}\E\left[ |f_\theta(X)-Y|^2\right]$.

An estimator for the minimal population mean squared error is given by the minimal empirical error on some sample $(X_1,Y_1),\ldots,(X_n,Y_n)\sim P$. Evidently, the consistency of this estimation corresponds to the convergence of the empirical process
\begin{align*}
	\Delta(\Ff)=\sup_{f\in\Ff} \Big| \E\left[ |f(X)-Y|^2 \right]- \frac{1}{n}\sum_{i=1}^{n}|f(X_i)-Y_i|^2 \Big|
\end{align*}
in probability, \ie, $\Delta(\Ff)\overset{n\to\infty}{\longrightarrow}_p 0$. The inner part of the above supremum is generally referred to as the \emph{generalization error}.  

By analyzing $\Delta(\Ff)$, one can identify conditions under which neural networks reach generalization ability. For binary-valued function classes, it is well-known that (under mild assumptions) $\Delta(\Ff)\overset{n\to\infty}{\longrightarrow}_p 0$ iff the function class has a finite VC-dimension \citep{wasserman_all_2004}. However, this finding is not directly applicable to networks with real-valued output as in our study. Instead, an analysis with real-valued output can be performed by introducing the more general concept of pseudo-dimension. An extensive overview over the corresponding results is given by \citet{anthony_neural_2009,vidyasagar_learning_2003}. We summarize concepts relevant to this work in the following.

\begin{definition}[Pseudo-dimension]
	A set $\{X_1,\ldots,X_n\}\subseteq\Xx$ is \emph{pseudo-shattered} by $\Ff$ if there is a shift-vector $r\in\R^n$ such that each binary combination $b\in\{0,1\}^n$ is realized by a function $f_b\in\Ff$, \ie, $\indicator{\geq 0}(f_b(x_i)-r_i)=b_i$ for $i=1,\ldots,n$. The cardinality of the largest pseudo-shattered subset of $\Xx$ is called the \emph{pseudo-dimension} $\mathrm{Pdim}(\Ff)$.
\end{definition}

\textbf{Sample complexity of learning algorithms.} Throughout this paper, we make the assumption of a sufficiently precise sample error minimization~(SEM) algorithm, \ie, a mapping which chooses a function $f\in\Ff$ for each finite sample such that the empirical error of $f$ is within $\varepsilon$ of the minimal empirical error over the whole function class. Given a function class with finite pseudo-dimension, a SEM algorithm directly implies the desired convergence of the empirical process. More precisely, it defines a selection rule $L:Z\mapsto f_{L_Z}\in\Ff$ for samples $Z=((X_1,Y_1),\ldots,(X_m,Y_m))$ such that for each $\varepsilon,\delta>0$, there is a positive integer $M(\varepsilon,\delta)$ such that 
$$
	\mathbb{P}^m\left(\E\left[|f_{L_Z}(X)-Y|^2\right]-\E\left[|f_{\theta^\ast}(X)-Y|^2\right]>\varepsilon\right)<\delta
$$
for each sample of length $m\geq M(\varepsilon,\delta)$. A function class (or neural network) with this property is called \emph{learnable}, the selection rule $L$ is a \emph{learning algorithm}, and the convergence rate $M_L(\varepsilon,\delta)$ is referred to as its \emph{sample complexity}.\footnote{If $M_L(\varepsilon,\delta)$ is polynomial, the function class is also called \emph{PAC-learnable}.}

The sample complexity defines a rate of consistency of the underlying error estimation. Given a population prediction error $\varepsilon$ and a desired confidence $\delta$, we can guarantee that a sample of length $M_L(\varepsilon,\delta)$ is large enough in order to find -- with high probability -- a network configuration with population error close to optimal. It only requires a bound on the network capacity measured by its pseudo-dimension. In fact, such bound can be used to directly infer a constructive sample complexity bound.

\begin{theorem}[Sample complexity of learning algorithms]
 \label{thm:sample_complexity}
	If $\Ff$ has finite pseudo-dimension, a SEM algorithm for $\Ff$ implies a learning algorithm with sample complexity
	$$
		M_L(\varepsilon,\delta)\leq \frac{128}{\varepsilon^2} \left[ 2\,\mathrm{Pdim}(\Ff)\ln\left(\frac{34}{\varepsilon}\right)+\ln\left(\frac{16}{\delta}\right) \right].
	$$
	\begin{proof}
		See \citet{anthony_neural_2009}.
	\end{proof}
\end{theorem}

Explicit bounds on the pseudo-dimension of neural networks are scarce. \citet{anthony_neural_2009} give few examples for feed-forward neural networks, however, a derivation for real-valued RNNs is still lacking.

\subsection{The upper bound for RNN sample complexity}

In the following, we derive explicit sample complexity bounds for learning real-valued RNNs. This is achieved by first providing a bound to their pseudo-dimension for the first time and, based on this, obtaining the actual sample complexity bound. Detailed proofs are reported in the supplementary materials. Upper bounds for the sample complexity are reported for both RNNs with a single recurrent layer and, further, multi-layer RNNs. In all derivations, we study RNNs with ReLU activation functions as these are common in machine learning practice. 

\subsubsection{Sample complexity bound for single-layer RNNs}
  
\begin{theorem}[Sample complexity bound for single-layer RNNs]
	\label{thm:mainsinglelayer}
	A recurrent neural network with
	\begin{enumerate}[noitemsep,topsep=0pt]
		\itemsep0em
		\item[(1)] a single recurrent layer of width $a$,
		\item[(2)] rectified linear units,
		\item[(3)] input of maximal length $b$, and 
		\item[(4)] one real-valued output unit
	\end{enumerate}
	is learnable with sample complexity that is bounded by
	\begin{align*}
	&M_L(\varepsilon,\delta) \leq  \frac{128}{\varepsilon^2} \Bigg[ \ln\left(\frac{16}{\delta}\right)+\ln\left(\frac{34}{\varepsilon}\right)\\
	&\times  4(a^2+3a+3)(2b (2a^2+4a)+4a+10+\log_2(8e)) \Bigg].
	\end{align*}
	\begin{proof}
		This first requires a novel bound to $\mathrm{Pdim}(\Ff)$ for such recurrent neural networks and, using \Cref{thm:sample_complexity}, one can eventually obtain the above bound for $M_L(\varepsilon,\delta)$; see supplementary materials for details.
	\end{proof}
\end{theorem}

The previous theorem has important implications: (1)~A population prediction error of $\varepsilon$ can be obtained with at most $\tilde{\mathcal{O}}\left(a^4b/\varepsilon^2\right)$ samples. Note that the $\tilde{\mathcal{O}}(\cdot)$ notation indicates that poly-logarithmic factors are ignored. (2)~As another insight, the number of required samples grows linearly with the maximum length of the input vector. (3)~The sample complexity grows at most polynomially with the number of recurrent units.

\subsubsection{Sample complexity bound for multi-layer RNNs}

Complex input such as sequential data or even graphs is often better processed when choosing multiple recurrent layers, as this allows to obtain functional representations that are more sophisticated. Hence, we extend our previous theoretical results to multi-layer ReLU RNNs. 

\begin{theorem}[Sample complexity bound for multi-layer RNNs]
	\label{thm:mainmulti-layer}
	A recurrent neural network with
	\begin{enumerate}[noitemsep,topsep=0pt]
		\itemsep0em
		\item[(1)] $d$ recurrent layers of widths $a_1,\ldots,a_d$,
		\item[(2)] rectified linear units,
		\item[(3)] input of maximal length $b$, and 
		\item[(4)] one real-valued output unit
	\end{enumerate}
	is learnable with a sample complexity that is bounded by
	\begin{align*}
	&M_L(\varepsilon,\delta) \leq  \frac{128}{\varepsilon^2} \Bigg[\ln\left(\frac{16}{\delta}\right)+\ln\left(\frac{34}{\varepsilon}\right)\\
	&\times  4\left(\sum_{i=1}^{d}(a_i^2+2a_i)+a_d+3\right)\Bigg(2b(2a_1^2+4a_1)\\
	&+\sum_{i=1}^{d-1}(a_i(4a_{i+1}^2+8a_{i+1}))+4a_d+10+\log_2(8e)\Bigg)\Bigg].
	\end{align*}
	\begin{proof}
		Again, we derive novel bound to $\mathrm{Pdim}(\Ff)$ for RNNs, based on which we can also bound $M_L(\varepsilon,\delta)$; see supplementary materials.
	\end{proof}
\end{theorem}

Even when choosing multiple recurrent layers, the polynomial nature of the sample complexity is maintained. For this, let $a_{\max} = \max \{ a_1, \ldots, a_d \}$ denote the largest width among all recurrent layers. Then, a RNN requires at most $\tilde{\mathcal{O}}(d^2a_{\max}^4b/\varepsilon^2)$ samples in order to learn a close-to-optimal configuration. In machine learning practice, the first hidden layer is commonly chosen to be the widest. In keeping with that, the sample complexity of multi-layer RNNs can be simplified to $\tilde{\mathcal{O}}(d^2a_1^4b/\varepsilon^2)$.

\section{Application of RNN sample complexity bounds to learning combinatorial graph problems}
\label{sec:eccn}

We now show how our theoretical results aid applications. For this reason, we leverage our sample complexity bounds in order to study the problem of learning combinatorial graph problems. This problem is particularly suited for RNNs: on the one hand, the recursive processing inside RNNs allows them to naturally handle graphs of varying size and, on the other hand, RNNs can numerically outperform feed-forward architectures as we shall see later in our numerical experiments. 

\subsection{Problem Statement}

Let $\mathcal{G}_n=\{0,1\}^{n\times n}$ denote the set of graphs with $n$ vertices represented by their adjacency matrices, and let $\mathcal{G}_{\leq n}=\mathcal{G}_1\cup\cdots\cup\mathcal{G}_n$ be the respective set of graphs with up to $n$ vertices. We consider undirected and unweighted graphs without self-loops, \ie, the entries of their adjacency matrices from $\mathcal{G}_n$ are symmetric with zero diagonal. 

We aim at learning an approximation to a graph problem $\eta: \mathcal{G}_{\leq n} \to \R$ with real-valued output based on a training data sample $\{(x_1, \eta(x_1)), (x_2, \eta(x_2)), \ldots\}$. We specifically decided upon this problem definition due to the fact that it resembles the general function classes $\Ff\subseteq \{ f: \Xx \to [0, 1] \}$ that are typically studied in the context of statistical learning theory. Potential constraints are not explicitly modeled but are inherently reflected in the mapping $\eta$ (\eg, infeasible input with invalid constraints would be mapped onto an arbitrarily large number). 

\subsection{Learning combinatorial graphs problems with single-layer RNNs}

\subsubsection{Single-layer RNN configuration}

We consider a RNN architecture with a single hidden layer and rectified linear units. Different from our earlier theorems, we introduce a size-adaptive approach, \ie, the number of recurrent units is directly set to $n$. This network is then fed with graphs with up to $n$ vertices but where, prior to that, the adjacency matrix of an $n$-vertex graph is mapped onto binary, one-dimensional vector $x=(x_1,\ldots,x_{n^2})$. In our RNN, the vector is processed by recursive calculations based on ReLUs, \ie,
$$
	h^i=(b+W^Th^{i-1}+U^Tx_i)_+ , \quad i = 1, \ldots, n^2,
$$
where $W\in\R^{n\times n}$ and $U\in\R^{1\times n}$ are the weight matrices of the recurrent layer and $b\in\R^{n}$ is the bias vector. Eventually, a single linear unit with weight $w^o\in\R^n$ and bias $b^o\in\R$ is used in order to obtain the final prediction via $x\mapsto w^{oT}h^{n^2}+b^o \in [0, 1]$. Statistically, boundedness of the outputs is required in order to ensure that the prediction errors do not grow arbitrarily large. We address this issue by placing several restrictions on the parameter space:
\begin{itemize}[topsep=0pt]
	\item Recurrent layer: $\lVert w_i\rVert_1\leq 0.25$ where $w_i$ is the $i$-th column of $W$, $\lVert U^T\rVert_\infty\leq 0.25 $, $\lVert b\rVert_\infty\leq 0.5$, and $\lVert h^0\rVert_\infty\leq 1$, for $i = 1, \ldots, n^2$.
	\item Output layer: $w^o\geq 0$, $\lVert w^o\rVert_1\leq 0.5$ and $0\leq b^o\leq 0.5$.
\end{itemize}
These restrictions ensure that $\text{image}(\text{RNN})\subseteq[0,1]$, which can be easily derived by iterative applications of H\"older's inequality.

\subsubsection{Sample complexity for the single-layer case}

Learning single-layer RNNs for the prediction of combinatorial graph problems has bounded sample complexity as follows.
\begin{theorem}
	\label{thm:graphsingle}
	The previous size-adaptive RNN with ReLU activation functions can be learned with sample complexity bounded by
	\begin{align*}
		&M_L(\varepsilon,\delta)\leq \frac{128}{\varepsilon^2}\Bigg[\ln\left(\frac{16}{\delta}\right)+\ln\left(\frac{34}{\varepsilon}\right)\\
		&4(n^2+4n+3)(4n^4+8n^3+4n+10+\log_2(8e))\Bigg] 
	\end{align*}
and thus $M_L(\varepsilon,\delta)\in\tilde{\mathcal{O}}(n^6/\varepsilon^2)$.
	\begin{proof}
	Based on H\"older's inequality, the restrictions on the RNN imply $\text{image}(\text{RNN})\subseteq[0,1]$ and, with Theorem~$\ref{thm:mainsinglelayer}$, the above sample complexity follows.
	\end{proof}
\end{theorem}

Accordingly, when learning an optimal network configuration with population prediction error $\varepsilon$ and certainty $\delta$, the sample complexity grows polynomially with the size of the graphs. More precisely, this growth is in $\mathcal{O}(n^6)$. 

\subsection{Learning combinatorial graphs problems with multi-layer RNNs}

\subsubsection{Multi-layer RNN configuration}

We extend the capacity of our previous single-layer RNN to a total of $d$ recurrent layers. Each of these layers operates on the last hidden state of the previous layer in a fully connected manner before transferring to a single real-valued linear output unit. We again choose $n$ units per layer in order to yield a size-adaptive architecture. Hence, the hidden states are computed recursively via\footnote{In order to distinguish indices, we print vector indices in a regular font and layer indices in bold.}
\begin{align*}
	\boldsymbol{h}^i_{\boldsymbol{1}}&=(b_{\boldsymbol{1}}+W_{\boldsymbol{1}}^T\boldsymbol{h}_{\boldsymbol{1}}^{i-1}+U_{\boldsymbol{1}}^Tx_i)_+\text{ for } i=1,\ldots,n^2,\\
	\boldsymbol{h}^i_{\boldsymbol{2}}&=(b_{\boldsymbol{2}}+W_{\boldsymbol{2}}^T\boldsymbol{h}_{\boldsymbol{2}}^{i-1}+U_{\boldsymbol{2}}^T\boldsymbol{h}_{\boldsymbol{1}i}^{n^2})_+\text{ for } i=1,\ldots,n,\\
%	\boldsymbol{h}^i_{\boldsymbol{3}}&=(b_{\boldsymbol{3}}+W_{\boldsymbol{3}}^T\boldsymbol{h}_{\boldsymbol{3}}^{i-1}+U_{\boldsymbol{3}}^T\boldsymbol{h}_{\boldsymbol{2}i}^n)_+\text{ for } i=1,\ldots,n,\\
	\vdots\\
	\boldsymbol{h}^i_{\boldsymbol{d}}&=(b_{\boldsymbol{d}}+W_{\boldsymbol{d}}^T\boldsymbol{h}_{\boldsymbol{d}}^{i-1}+U_{\boldsymbol{d}}^T\boldsymbol{h}_{\boldsymbol{d-1}i}^n)_+\text{ for } i=1,\ldots,n,
\end{align*}
for each hidden layer $\boldsymbol{h_j}=\boldsymbol{h_1},\ldots,\boldsymbol{h_d}$ subject to appropriate weight matrices $W_{\boldsymbol{j}}, U_{\boldsymbol{j}}$, a bias vector $b_{\boldsymbol{j}}$, and rectifier activation function $(\cdot)_+$. Again, a final prediction is obtained by $x\mapsto w^{oT}\boldsymbol{h_d}^n+b^o$ where $w^o$ and $b^o$ are the parameters of the output unit. 

\subsubsection{Sample complexity for the multi-layer case}

With appropriate restrictions to the parameters, the mulit-layer RNN defines a class of functions $\{f_d:\mathcal{G}_{\leq n}\to [0,1]\}$. It obtains the following bounded sample complexity for the prediction of combinatorial graph problems. 

\begin{theorem}
	\label{thm:graphmulti}
	The previous size-adaptive multi-layer RNN can be learned with sample complexity bounded by
	\begin{align*}
		&M_L(\varepsilon,\delta)\leq\frac{128}{\varepsilon^2}\Bigg[\ln\left(\frac{16}{\delta}\right)+\ln\left(\frac{34}{\varepsilon}\right)\\
		&4(dn^2+(2d+1)n+3)(4n^4+(8+4(d-1))n^3\\
		&+8(d-1)n^2+4n+10+\log_2(8e))\Bigg] \in \tilde{\mathcal{O}}(d^2n^6/\varepsilon^2).
	\end{align*}
	\begin{proof}
		Analogous to \Cref{thm:graphsingle}.
	\end{proof}
\end{theorem}

Based on \Cref{thm:graphmulti}, we supplement our earlier findings in multiple ways. (1)~At most $\tilde{\mathcal{O}}(d^2n^6/\varepsilon^2)$ samples are required in order to reach a population prediction error of $\varepsilon$. 
(2)~The inherent complexity of graph problems -- especially for large graphs -- might benefit from multiple recurrent layers in order to attain a favorable approximation performance. Our analysis shows that, for a fixed problem size $n$ and prediction error $\varepsilon$, the addition of further recurrent layers merely increases the sample size to $\tilde{\mathcal{O}}(d^2)$. This increase in the sample complexity appears fairly small (\eg, when compared with $n$) and thus renders a higher depth of network feasible from a theoretical perspective. (3)~We chose a problem-size-adaptive approach to the number of recurrent layers (\ie, $d=n$) and, despite that, the sample complexity remains polynomial in the maximal graph size $n$. It is thus plausible that a model with a sufficient approximation performance can be retrieved eventually.

\subsection{Discussion of findings}

\textbf{Comparison of sample complexity to input space.} As a key finding, this paper for the first time proves a polynomial growth of the sample complexity for training of RNNs as heuristic solvers for combinatorial graph problems. The same theoretical considerations are used to generalize to multi-layer architectures, while maintaining a polynomial rate of consistency.
This is a remarkable result, as both the complexity of a combinatorial problem and the number of distinct graphs typically grow exponentially with the graph size, thus rendering direct calculations computationally intractable for many combinatorial graph problems. The different growth rates suggest that a break-even point $n = n^\ast$ exists for which the sample complexity is smaller than the input space of $\lvert\mathcal{G}_{\leq n}\rvert=\sum_{i=1}^{n}2^{(i^2-i)/{2}} \in \mathcal{O}(2^{n^2})$ graphs. As an example, we assume $\varepsilon = \delta = 0.1$ in the following. Then, for $n = 10$, only 5.82\,\% of all graphs in $\mathcal{G}_{\leq n}$ are needed to satisfy the sample complexity bound of the proposed single recurrent layer RNN model, whereas, for $n = 11$, the required share of training samples drops to $0.963 \times 10^{-3}$\,\%. Hence, the number of required training samples for finding a learning algorithm is considerably lower than the dimension of the input space. 

\textbf{Learning given predefined population prediction error.} As a powerful implication of \Cref{thm:mainsinglelayer,thm:mainmulti-layer}, the above theoretical results even allow for straightforward construction of learning algorithms with a predetermined statistical accuracy $\varepsilon$. Such algorithm can be derived based on the assumed SEM algorithm $\mathcal{A}$ by setting the learning algorithm to $L(z)=\mathcal{A}(z,8/(3\sqrt{m}))$ for a sufficiently large sample with $m$ entries. For details on obtaining a desired population prediction error $\varepsilon$, we refer to \citep{anthony_neural_2009}.

As we shall also confirm in our numerical experiments, the polynomial rate renders our approach competitive to other heuristic solvers in the operations research literature. This also reveals a strength of RNNs for such undertakings, as RNNs are naturally flexible in learning from input of variable size. 

\section{Numerical experiments for prediction performance}
\label{sec:experiments}

This section amends the theoretical findings through numerical experiments the following reasons: (1)~Our notion of sample complexity defines a rate of statistical consistency and is thus detached from the approximation performance of the heuristic solver. Our experimental results show that RNNs indeed comes with the required approximation capacity to compete against and outperform traditional heuristic approaches. (2)~In our experiments, RNNs appear superior over feed-forward networks. This justifies our effort in bounding the sample complexity of RNNs. (3)~In practical settings the effective sample complexity, \ie, the number of samples effectively required to reach some generalization properties, could be considerably smaller than the upper bound derived above. This stems from the fact that our theoretical considerations are distribution-independent and, hence, reflect some kind of worst-case scenario.

\subsection{Example problem: prediction of edge clique cover numbers}

We have specifically chosen the edge clique cover number (ECCN) problem 
 for demonstrating the performance of our RNN-based approach. The reasons are three-fold: (1)~ECCN is NP-hard and thus computationally challenging. (2)~It is relevant to a variety of practical applications, including computational geometry \citep{agarwal_can_1994}, compiler optimization \citep{rajagopalan_handling_2000}, computational statistics \citep{gramm_algorithms_2007,piepho_algorithm_2004}, protein interaction networks \citep{blanchette_clique_2012}, and real-world network analysis \citep{guillaume_bipartite_2004}. (3)~Known heuristics are scarce, with the Kellerman heuristic being the notable exception \citep{kellerman_determination_1973,kou_covering_1978}.

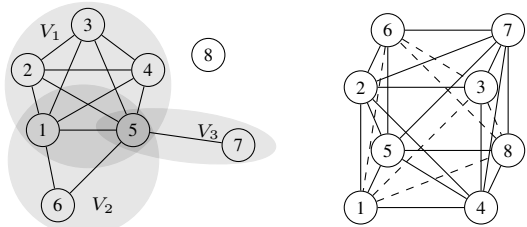
\begin{figure}
	\centering
		\scriptsize
		\begin{minipage}{0.2\textwidth}
			\raggedright
		\begin{tikzpicture}[scale=.2]
		\begin{scope}[transparency group]
		\begin{scope}[blend mode=multiply]
		%shaded
		\filldraw[black!10!white,rotate=-7] (9.5,0.75) ellipse (6cm and 1.7cm);
		\filldraw[black!10!white] (3,3) circle (5.5cm);
		\filldraw[black!10!white] (2.7,-2) circle (5cm);
		\end{scope}
		\end{scope}
		%environment names
		\node (V1) at (0.5,6.5) {$V_1$};
		\node (V2) at (4,-5) {$V_2$};
		\node (V3) at (11,-0.1) {$V_3$};
		%vertices
		\node[shape=circle, draw=black] (1) at (0,0) {1};
		\node[shape=circle, draw=black] (2) at (-1,4) {2};
		\node[shape=circle, draw=black] (3) at (3,7) {3};
		\node[shape=circle, draw=black] (4) at (7,4) {4};
		\node[shape=circle, draw=black] (5) at (6,0) {5};
		\node[shape=circle, draw=black] (6) at (1,-5) {6};
		\node[shape=circle, draw=black] (7) at (13,-1) {7};
		\node[shape=circle, draw=black] (8) at (11,5) {8};
		%edges	
		\path (1) edge node[left] {} (2);
		\path (2) edge node[left] {} (3);
		\path (3) edge node[left] {} (4);
		\path (4) edge node[left] {} (5);
		\path (5) edge node[left] {} (1);
		\path (1) edge node[left] {} (3);
		\path (1) edge node[left] {} (4);
		\path (2) edge node[left] {} (4);
		\path (2) edge node[left] {} (5);
		\path (3) edge node[left] {} (5);
		\path (1) edge node[left] {} (6);
		\path (6) edge node[left] {} (5);
		\path (5) edge node[left] {} (7);
		\end{tikzpicture}		
		\end{minipage}
		\begin{minipage}{0.2\textwidth}
			\raggedleft
		\begin{tikzpicture}[scale=.8]
		%vertices
		\node[shape=circle, draw=black] (1) at (0,0) {1};
		\node[shape=circle, draw=black] (2) at (0,2) {2};
		\node[shape=circle, draw=black] (3) at (2,2) {3};
		\node[shape=circle, draw=black] (4) at (2,0) {4};
		\node[shape=circle, draw=black] (5) at (0.45,0.95) {5};
		\node[shape=circle, draw=black] (6) at (0.45,2.95) {6};
		\node[shape=circle, draw=black] (7) at (2.45,2.95) {7};
		\node[shape=circle, draw=black] (8) at (2.45,0.95) {8};
		%edges	
		\draw[dashed] (1)--(8)--(6)--(1)--(3)--(8);
		\draw[dashed] (6)--(3);
		\draw (1)--(2)--(3)--(4)--(1);
		\draw (2)--(4)--(8)--(5)--(1);
		\draw (4)--(5)--(6)--(7)--(8);
		\draw (5)--(7)--(4);
		\draw (3)--(7)--(2);
		\draw (6)--(2)--(5);
		\end{tikzpicture}
		\end{minipage}
	\caption{
		Visualization of an example graph with cliques $V_1=\{1,2,3,4,5\}$,$V_2=\{1,5,6\}$ and $V_3=\{5,7\}$ forming a minimal edge clique cover. No two cliques can cover all edges; hence, the ECCN of $G$ is $3$ (left). Die-shaped graph $G$ with 16 distinct maximal cliques. A straightforward edge clique cover can be obtained by choosing the 6 sides of the die as maximal cliques, but we see that the ECCN of $G$ is 5 with the choice of cliques $V_1=\set{1,3,6,8}$, $V_2=\set{1,2,4,5}$, $V_3=\set{2,3,4,7}$, $V_4=\set{2,5,6,7}$ and $V_5=\set{4,5,7,8}$ (right).}
	\label{fig:graph}
\end{figure}

An edge clique cover number refers to the minimum number of cliques, \ie, fully connected sub-graphs, required to cover all edges (see \Cref{fig:graph}).

\begin{definition}[Edge clique cover number]
	Let $G=(V,E)$ be an undirected graph and $\Vv=\set{V_{1},\ldots,V_{l}}$ a set of cliques. We denote with $G_i=(V_i,E_i)$ the subgraph induced by $V_i$ for $i=1,\ldots,l$. Then, $\Vv$ is called an edge clique cover of $G$, if
	$
	E_1\cup\ldots\cup E_l = E.
	$
	If $\Vv$ is of minimal cardinality with this property, then $l$ gives the ECCN of $G$.
\end{definition}

\subsection{Computational setup}

The goal is to shed light into the characteristics of our neural network approach numerically. We utilize the following networks for our comparisons: (1)~a multi-layered feed-forward network~(FNN) with ReLUs that receives an adjacency matrix as input, (2)~a recurrent neural network as in our derivations, and (3)~a long short term memory network~(LSTM) as a more complex architecture with a cell structure that finds widespread application in practice. Both RNN and LSTM process the adjacency matrix sequentially. Analogous to our proofs, all activation functions are set to the rectifier function $(\cdot)_+$. In addition, we compare our methods against the state-of-the-art heuristic (Kellerman heuristic) proposed by \citep{kellerman_determination_1973,kou_covering_1978}. See supplementary materials for exact training details.

Our experiments draw upon input given by random Erd\"os-Renyi graphs (\ie, edges are formed with probability $p$) based on different scenarios: sparse ($p=0.1$), medium dense ($p=0.5$), and dense graphs ($p=0.9$), as well as a mixture thereof. They contain graphs of different sizes, ranging between $10$ to $18$ nodes in order to validate the flexibility of prediction performance in the presence of varying $n$. The maximum size was limited to 18 nodes, since the labeling requires solving the NP-hard ECCN problems, which is computationally intractable for larger values. %See supplementary material for details.

\subsection{Performance across training sample size}

\begin{figure*}[htb]
	\footnotesize
	\makebox[\textwidth]{
		\begin{tabular}{cccc}
			(a)~dense graphs & (b)~medium dense graphs & (c)~sparse graphs & (d)~mixed graphs \\
			\includegraphics[width=.22\textwidth]{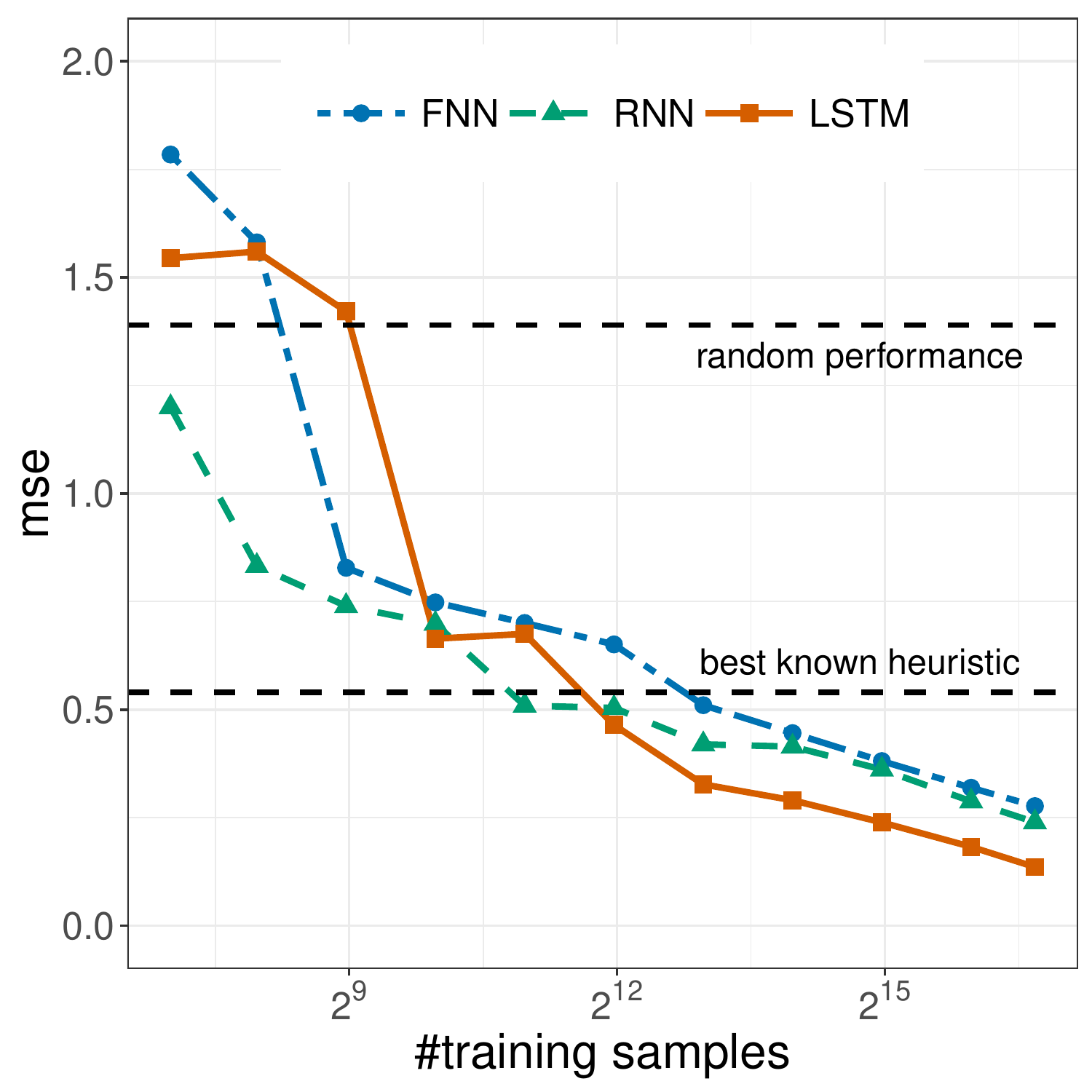}  &
			\includegraphics[width=.22\textwidth]{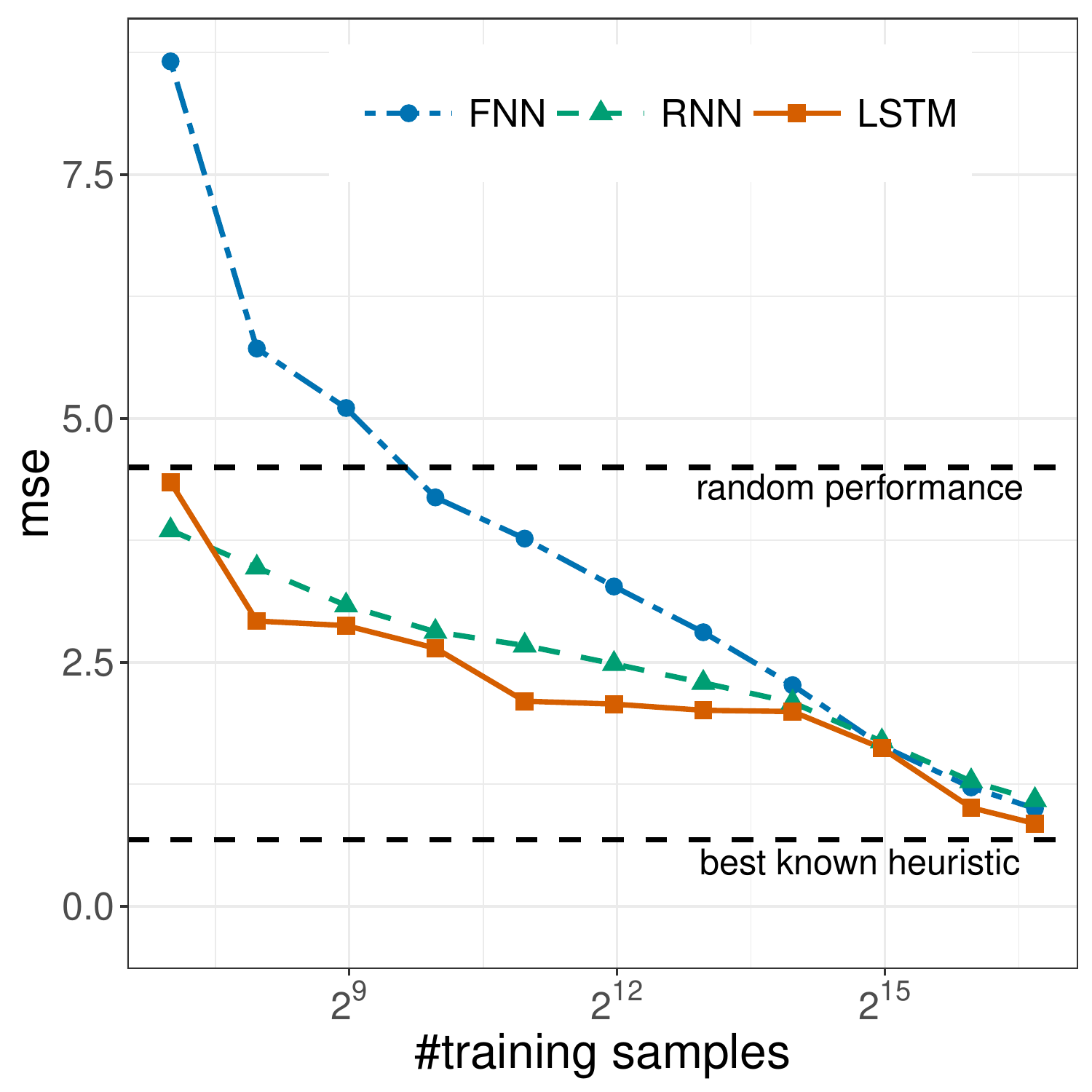} &  \includegraphics[width=.22\textwidth]{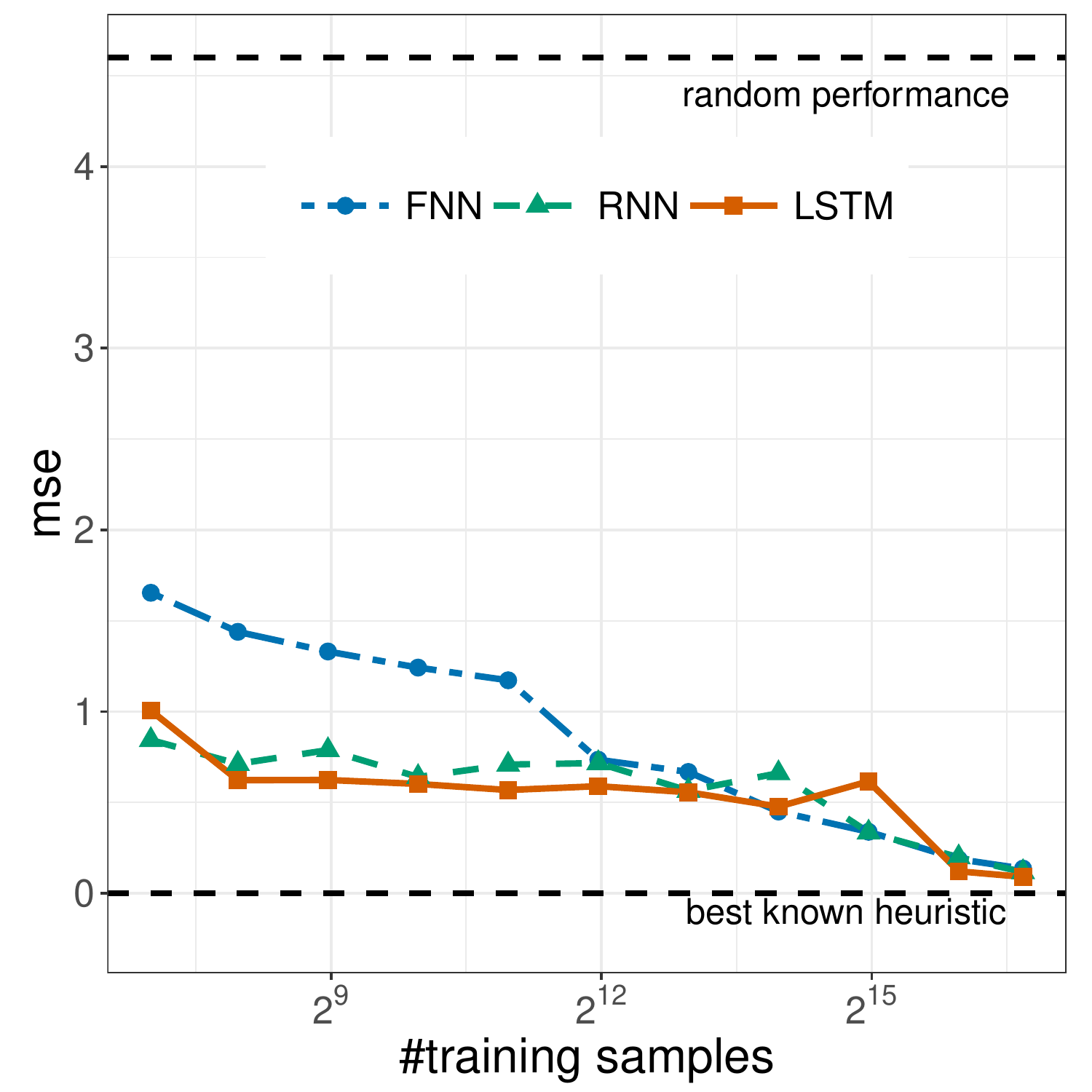} &
			\includegraphics[width=.22\textwidth]{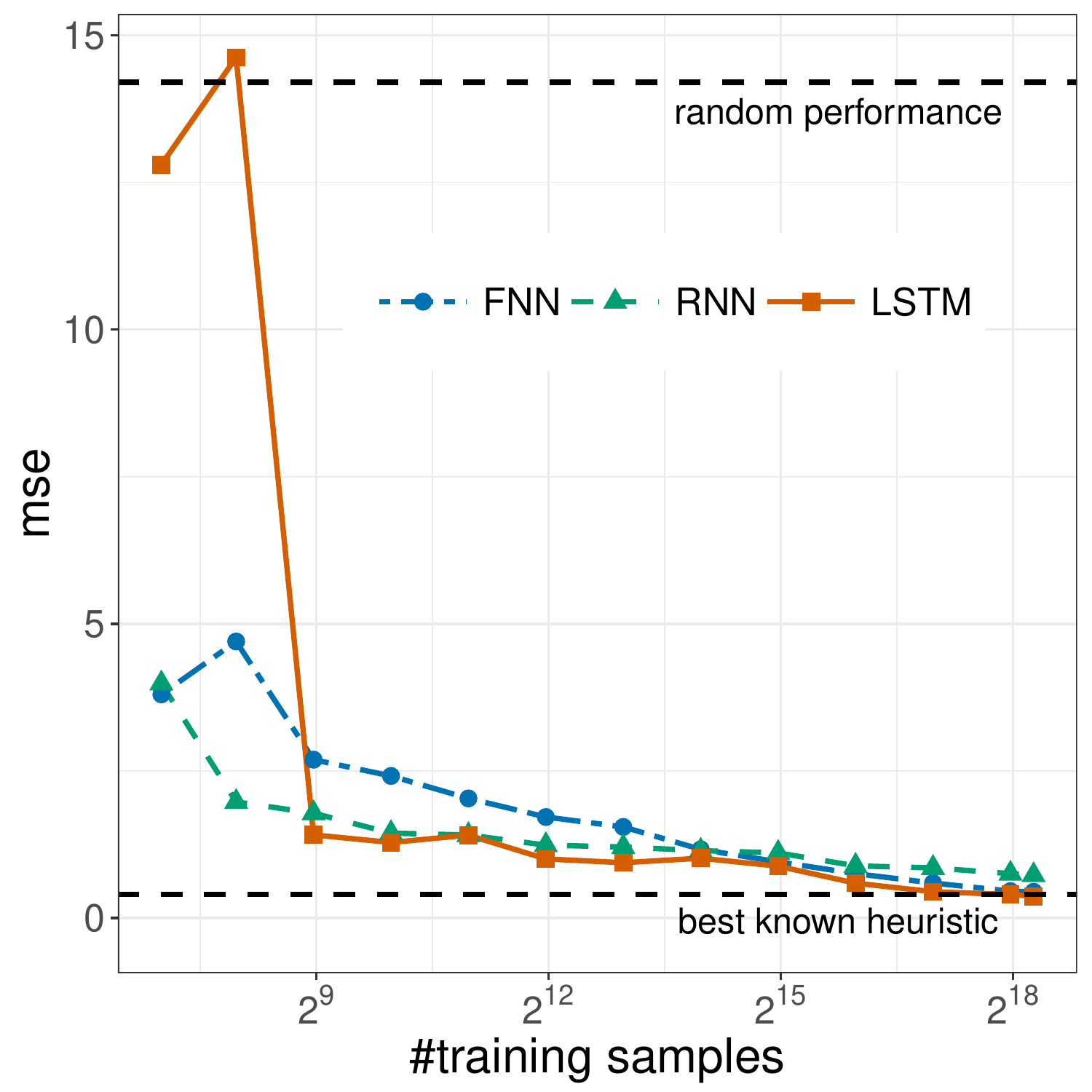} \\
		\end{tabular}
	}
	\caption{Out-of-sample performance of learning predictions of the edge clique cover number. The mean squared error~(mse) is reported as a function of the number of training samples (log-scale). With sufficient training samples, RNN predictions can successfully outperform a na{\"i}ve baseline (\ie, majority vote) and, in some scenarios, even the state-of-the-art heuristic (Kellerman heuristic).}
	\label{fig:samplesizecomplexity}
\end{figure*}

We numerically evaluate how the number of training samples affects the out-of-sample performance in predicting solutions of the combinatorial problem. We thus plot the mean squared error as a function of the number of training samples (log-scale) in \Cref{fig:samplesizecomplexity}. The findings justify our derivations for RNNs (as opposed to feed-forward networks) as these appear largely superior. With sufficient training data, neural learning over graphs is either on par with the Kellerman heuristic and often even outperforms it. Notably, a reasonable performance can often be achieved with as little as 4,000 training samples. This contributes to our claim of inherent approximation capabilities of the proposed neural network models. In fact, a satisfactory performance in practical applications seems to require considerably fewer training samples than the derived upper bound to the sample complexity suggests. 

\begin{figure*}[htbp]
	\makebox[\textwidth]{
	\begin{minipage}{0.22\textwidth}	
		\includegraphics[width=1\textwidth]{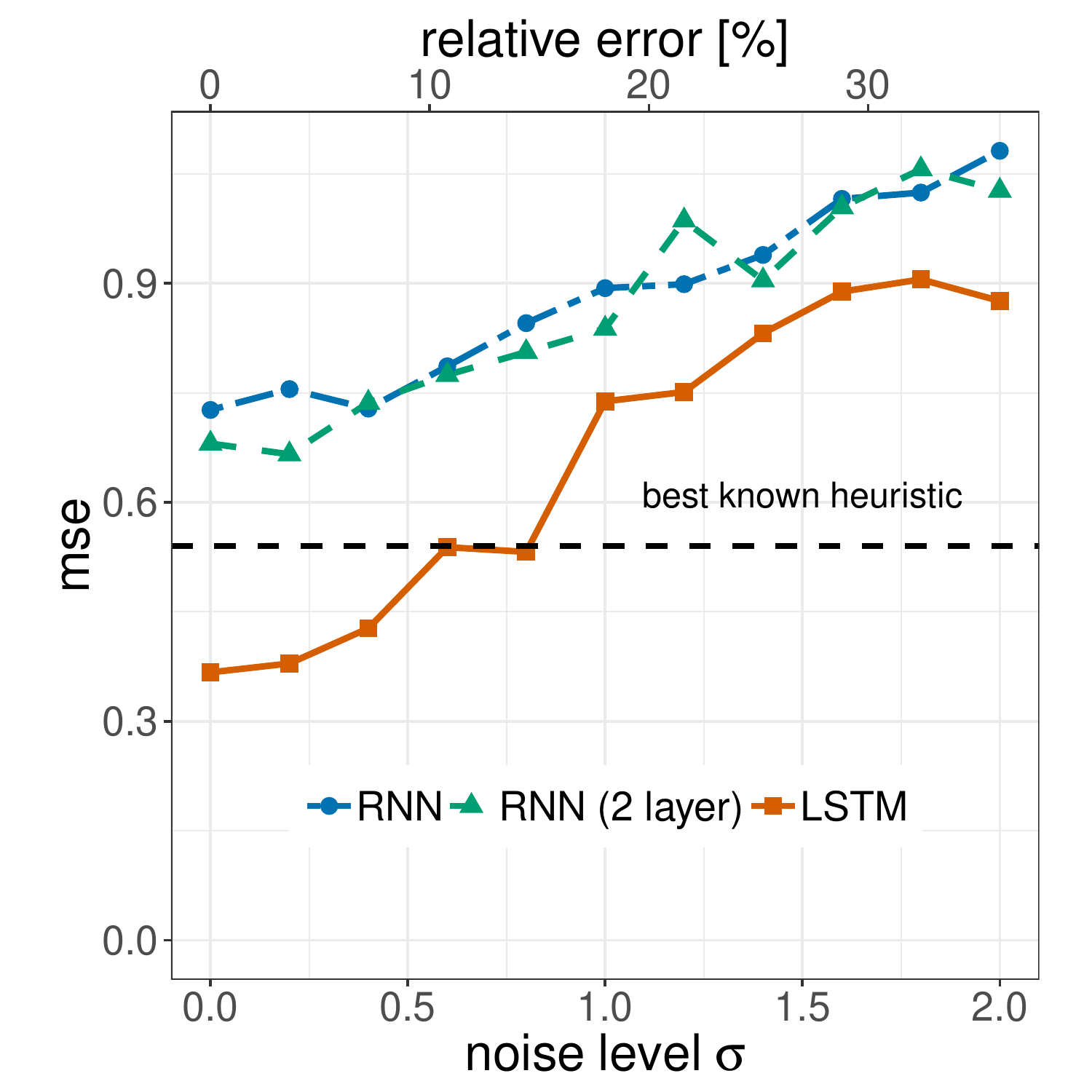}
	\end{minipage}
	%\begin{minipage}{0.00\textwidth}~\end{minipage}
	\begin{minipage}{0.22\textwidth}	
		\includegraphics[width=1\textwidth]{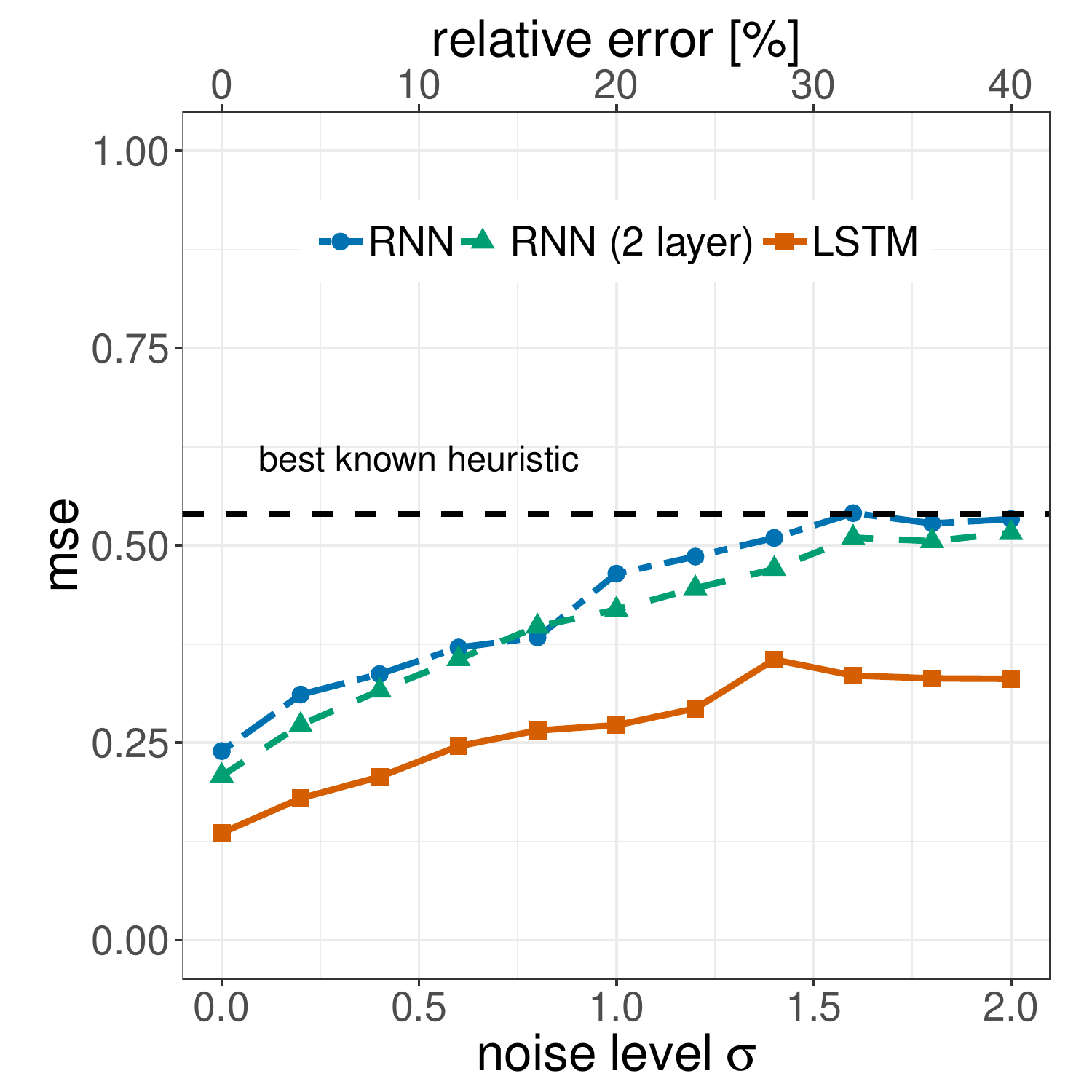}
	\end{minipage}
	%\begin{minipage}{0.00\textwidth}~\end{minipage}
	\begin{minipage}{0.5\textwidth}		
		\caption{
			Out-of-sample performance under noise, \ie, when the training data is perturbed by Gaussian noise with different standard deviations $\sigma$. Left: scenario with a mixture of sparse and dense graphs; right: purely dense graphs. Even when adding a relative error of 10\,\% or more, the predictions can still outperform the state-of-the-art heuristic. The single-layer RNN is outperformed slightly by the multi-layer RNN with two recurrent layers. 
		}
		\label{fig:graph_noise}
	\end{minipage}
}
\end{figure*}

\subsection{Sensitivity of learning under noise}

\Cref{fig:graph_noise} further establishes that neural networks are effective when being trained with noisy input. For this purpose, we perturb the original labels with Gaussian noise, \ie, $y'_\text{train} = y_\text{train} + \epsilon$ with $\epsilon \sim N(0,\sigma)$. Here we draw upon dense graphs as these are particularly challenging for the state-of-the-art heuristic. We observe that $d > 1$ layers outperforms the single-layer RNN. Accordingly, our approach appears robust, even with high levels of noise (\eg, relative errors of 10\,\% or more). This benefits real-world use cases, since it is often straightforward to construct training data with near-optimal solutions in a bottom-up fashion; for instance, one can easily construct graphs with a predefined ECCN through dynamic programming, whereas the opposite direction of inferring the ground-truth ECCN for a random graph is NP-hard.

\section{Conclusion}
\label{sec:conclusions}

Sample complexity bounds have been derived for various neural network architectures, such as, \eg, binary feed-foward neural networks 
 and convolutional neural networks. 
 Yet we are not aware of any previous work that is concerned with real-valued recurrent neural networks and, to overcome this research gap, we thus bound the sample complexity of RNNs. Our analysis shows that single-layer RNNs with $a$ ReLUs and inputs of length at most $b$ can be learned with $\tilde{\mathcal{O}}(a^4b/\varepsilon^2)$ samples in order to attain a population prediction error of $\varepsilon$. $d$-layer RNNs (where $a_{\max}$ denotes the number of units in the widest recurrent layer) require at most $\tilde{\mathcal{O}}(d^2a_{\max}^4b/\varepsilon^2)$ samples.

Bounding the sample complexity of RNNs enables interesting applications, as demonstrated in our work when learning combinatorial problems over graphs. While the numerical performance of neural networks has been previously studied for this application, theoretical guarantees remain scarce. As a remedy, this study presents the first effort at deriving sample complexity bounds for approximating combinatorial graph problems. For this application, a size-adaptive RNN obtains a sample complexity bounded by $\tilde{\mathcal{O}}(n^6/\varepsilon^2)$, which grows polynomially with the size of the graph. The polynomial sample complexity renders neural learning for this application competitive against traditional heuristic solvers. However, neural learning is problem-independent and thus eliminates the need for manual and non-trivial derivations of such solution heuristics. Our work is further accompanied with numerical experiments that study the approximation performance: based on the example of the NP-hard edge clique cover number problem, neural learning demonstrates competitive results and frequently outperforms the best-known heuristic. 

\FloatBarrier

\bibliography{literature}
\bibliographystyle{icml2019}

\appendix
\newpage

\section{Related work on neural learning of combinatorial graph problems}
\label{sec:relatedwork}

\textbf{Combinatorial problems.} Combinatorial problems over finite graph structures constitutes a fundamental problem at the intersection of applied mathematics and computer science \citep{korte_combinatorial_2018}.
Examples include the traveling salesman problem, the shortest path problem, and the edge clique cover number. These combinatorial graph problems arise directly from real-world applications, such as in transportation
\citep[\eg,][]{lenstra_simple_1975,park_school_2010}, communication networks \citep[\eg,][]{oliveira_survey_2005,chen_approximability_2009}, or scheduling \citep[\eg,][]{crama_combinatorial_1997}.

The use of neural networks has been suggested as heuristic for combinatorial graph problems \citep{aiyer_theoretical_1990,hopfield_neural_1985}. These network learn approximations to the actual task from labeled data. The intuition is that neural networks identify a mapping between pairs of graphs and solutions that generalizes to unseen input, thereby achieving a generalizable approximation to the true solution. We refer to \citet{smith_neural_1999,syed_neural_2013} for extensive overviews on the use of neural network models for combinatorial problems. The resulting solver promises an efficient computation scheme for combinatorial problems with clear benefits: it requires only a fixed evaluation cost when being applied to unseen data, and obtains solution schemes, even when any form of heuristic is unknown.

\textbf{Neural networks for combinatorial optimization.} Extensive research has applied neural networks to tasks from (combinatorial) optimization; see the overviews in \citep{smith_neural_1999,syed_neural_2013}, including various tasks (\eg, transportation and scheduling \citep{gupta_selecting_2000}) and ranging from simple feed-forward networks to differential neural computers \citep{graves_hybrid_2016}. However, the previous works rely purely on numerical demonstration and thus lack a rigorous theoretical understanding of the underlying estimation. 

\textbf{Neural search process with energy function.} Time-dependent networks with energy functions, such as Hopfield networks, have been utilized for solving the traveling salesman problem by reformulating the objective as an energy function that is then optimized \citep{hopfield_neural_1985}. This has been proven to converge to a fixed point in an input space with a projected traveling salesman problem \citep{aiyer_theoretical_1990}. However, in order to apply this approach to other optimization problems, one must carefully adapt the energy function to the specific task \citep{wen_review_2009}. Among their further drawbacks, these approaches sometimes only converge when the network dimension grows with the complexity of the solution (instead of the input size \citep{wen_review_2009}), entail inferior runtimes \citep[\eg,][]{smith_neural_1999}, and cannot yield single-shoot predictions but rather lead to a search process similar to reinforcement learning. 

\textbf{Iterative solvers through trial and error learning.} Deep reinforcement learning techniques have been shown to yield heuristic solvers for combinatorial graph problems \citep{bello_neural_2017,dai_learning_2017,nazari_deep_2018}. In \citet{bello_neural_2017}, the authors iteratively identify greedy solutions through a pre-trained pointer-network within a reinforcement framework. \citet{dai_learning_2017} draw upon model-free Q-learning to greedily add vertices or nodes in order to obtain a satisfactory solution in bottom-up fashion. The latter work is extended by \citet{nazari_deep_2018} to the vehicle routing problem, which is a dynamic generalization of the traveling salesperson problem.
This general approach of reinforcement learning follows a paradigm different from that of traditional neural network approaches, since its training process requires an interactive environment whereby the policy is continuously updated. As a result, the convergence process cannot a priori be foreseen and this thus yields computer programs with unclear runtimes.
As a consequence, both paradigms demand distinct theoretical guarantees: trial-and-error learning would -- in a first step -- need a proof of convergence \citep[cf.][]{singh_convergence_2000}, whereas we prove explicit polynomial bounds to the sample complexity of predicting solutions to combinatorial graph problems in a supervised manner.

\textbf{Synergies.} The above descriptions outline several differences between (i)~supervised learning of graph solutions and (ii)~trial-and-error learning, yet both paradigms can benefit from each other. On the one hand, predictions could be used to obtain a first, near-optimal estimate that is subsequently refined through a policy-based search. On the other hand, the search process in reinforcement learning could provide a data-driven, model-free approach to generating training samples for direct predictions of combinatorial graph problems. Here our experiments with regard to learning under noise hint that this might be a viable approach. Altogether, it would be interesting to see combinations of both approaches in the future. 

\section{Theoretical contributions}

\subsection{Proof technique and preliminaries}
\label{sec:skeleton}
In order to derive our sample complexity bounds, we draw upon empirical process theory, a traditional understanding of shattering, and a straightforward notion of neural network complexity based on the number of trainable parameters. Our proofs are composed of several steps which we will explain in this section.
Whereas the most important background is reviewed in the main paper, we refer to \cite{anthony_neural_2009} for a comprehensive overview on the mathematical basis.

\textbf{Sample complexity v.\,s. pseudo-dimension.} The basic idea of our proofs is to reduce questions of sample complexity to pseudo-dimension considerations. By Theorem~2.2 in the main paper, a real-valued neural network  $\Ff=\set{f_\theta:\Xx\to[0,1]:\theta\in\Theta}$ can be learned with sample complexity 
\begin{align*}
M_L(\varepsilon,\delta)\leq \frac{128}{\varepsilon^2}\left[2\text{ Pdim}(\Ff)\ln\left(\frac{34}{\varepsilon}\right)+\ln\left(\frac{16}{\delta}\right)\right],
\end{align*} 
if it has finite pseudo-dimension $\text{Pdim}(\Ff)<\infty$. We will therefore concentrate our efforts on the derivation of upper bounds for the pseudo-dimensions for recurrent ReLU networks.
However, there is only very little previous work concerned with techniques for direct bounding of pseudo-dimensions of neural networks. Instead, our proofs draw on the more explored notion of VC-dimension.

\textbf{Pseudo-dimension v.\,s. VC-dimension.}
With the aim of building on previous proof techniques for binary function classes, we relate the pseudo-dimension of a neural network to the VC-dimension of an extension of the network. In particular, we add a new thresholded linear output unit, which takes as input the original output and an additional real-valued input as comprised in the following lemma.

\begin{lemma}
	\label{lem:VCpseudo}
	Let $$\Hh=\set{\indicator{\geq 0}(f(x)+wy+b):f\in\Ff,w,b\in\R}$$ define a class of functions on $\Xx\times \R$, then it holds that $\text{Pdim}(\Ff)\leq \text{VCdim}(\Hh).$
	\begin{proof}
		See \citet{anthony_neural_2009}.
	\end{proof}
\end{lemma}

Now, upper bounding of the pseudo-dimension and, hence, upper bounding of the sample complexity boils down to VC-dimension arguments, which can be made on the basis of computational complexity.

\textbf{Concept classes and computational complexity.}
In some sense, the VC-dimension of a binary output network is a straightforward measure of its ability to distinguish between a number of inputs. Since this is a combinatorial question, it relates directly to the variety of computations the network can make.
In order to make use of this idea, we rely on the notion of concept classes \citep[\eg,][]{goldberg_bounding_1995}.

\begin{definition}[Concept class]
	A concept for a function $f_\theta\in\Ff$ is a subset $C_{f_\theta}\subset\Xx$ such that $x\in C_{f_\theta}$ iff $f_\theta(x)>0$. The family $\set{C_{f_\theta}:f_\theta\in\Ff}$ is called concept class.
\end{definition}

The key to leveraging the combinatorial nature of VC-dimension is then based on the derivation of the worst-case computational complexity an algorithmic procedure requires in order to decide whether an input $x\in\Xx$ belongs to some concept.
In order to derive this complexity, we count all arithmetic operations on $\R$, \ie $+,-,\times,/$, jumps conditioned on $<,\leq,>,\geq, =, \neq$, and outputs of ``true'' and ``false'' the network performs. 
This quantity evidently depends on the number of trainable network parameters $W$, and, since recurrent neural networks process inputs of variable size, also on the maximal input length $\tau$. Denoting the computational complexity $T=T(W,\tau)$, we can then conclude the following theorem.

\begin{theorem}
	\label{thm:vccomputationalcomplexity}
	$\text{VCdim}(\indicator{\geq 0}(\Ff))\leq 2W(2T+\log_2(8e))$.
	\begin{proof}
		See \citet{goldberg_bounding_1995}.
	\end{proof}
\end{theorem}

\subsection{Proof of Theorem~2.3}

For the single hidden layer model, we assume a width of $a$ recurrent rectified linear units and input sequences which are processed entry-wise. The ReLU RNN employs a single real-valued output unit.
Our first lemma counts the number of adjustable parameters of this architecture. 

\begin{lemma}
	\label{lem:numberparamssinglehiddenlayer}
	The described RNN has $W=a^2+3a+1$ trainable weights and biases.
	\begin{proof}
		In each time step, the RNN takes a one-dimensional entry of some input sequence and incorporates it into a $a$-dimensional hidden state by creating a weighted average of the previous hidden state and the new input component. Thus for some input sequence $x=(x_1,\ldots,x_\tau)$, input processing is based on recursive calculation of $$h^i=(b+W^Th^{i-1}+U^Tx_i)_+$$ with some weight matrices $W\in\R^{a\times a}$, $U\in\R^{1\times a}$ and a bias vector $b\in\R^{a}$. These parameters are shared throughout the time steps and, hence, add up to $a^2+2a$ adjustable one-dimensional weights and biases. From the last hidden state, a final prediction is generated by a linear unit, which calculates $w^{oT}h^\tau+b^{o}$ subject to a weight vector $w^{o}\in\R^a$ and bias $b^o\in\R$. In turn, the whole process relies on $W=a^2+3a+1$ adjustable parameters.
	\end{proof}
\end{lemma}

\begin{figure*}[h!]
	\centering
	\makebox[\textwidth]{
		\begin{minipage}{0.25\textwidth}
			\centering
			\includegraphics[scale = 0.15]{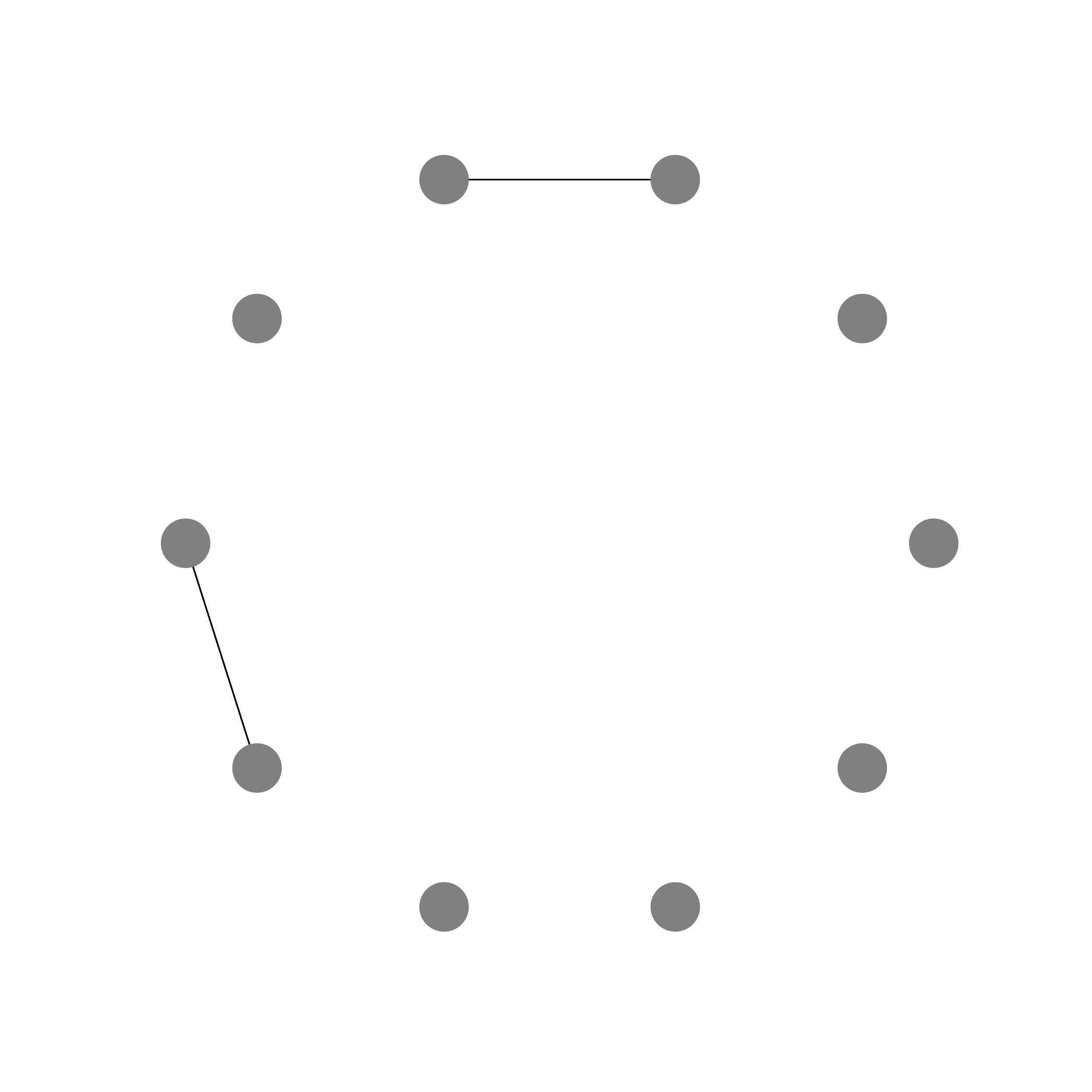}
		\end{minipage}
		\begin{minipage}{0.25\textwidth}
			\centering
			\includegraphics[scale = 0.15]{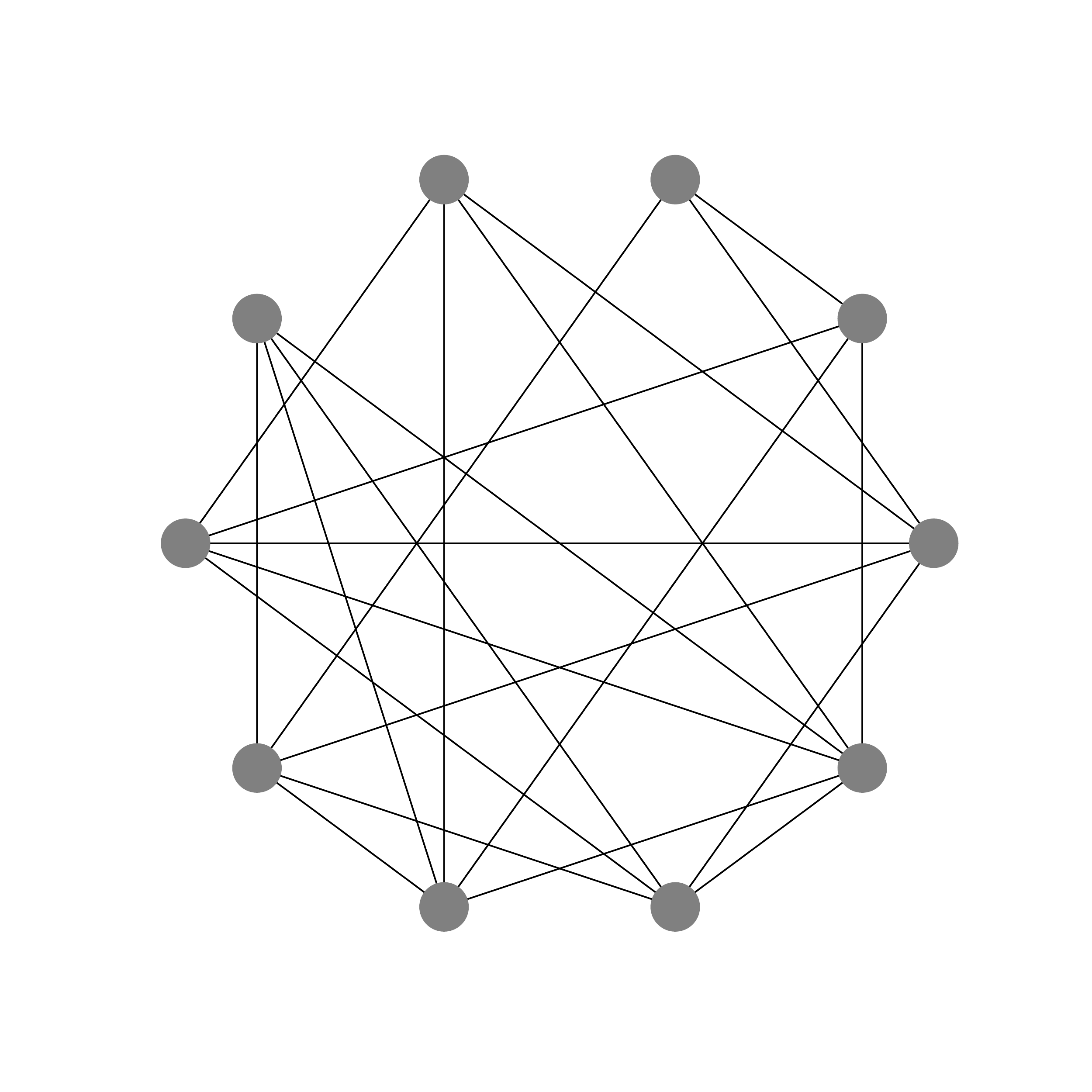}
		\end{minipage}
		\begin{minipage}{0.25\textwidth}
			\centering
			\includegraphics[scale = 0.15]{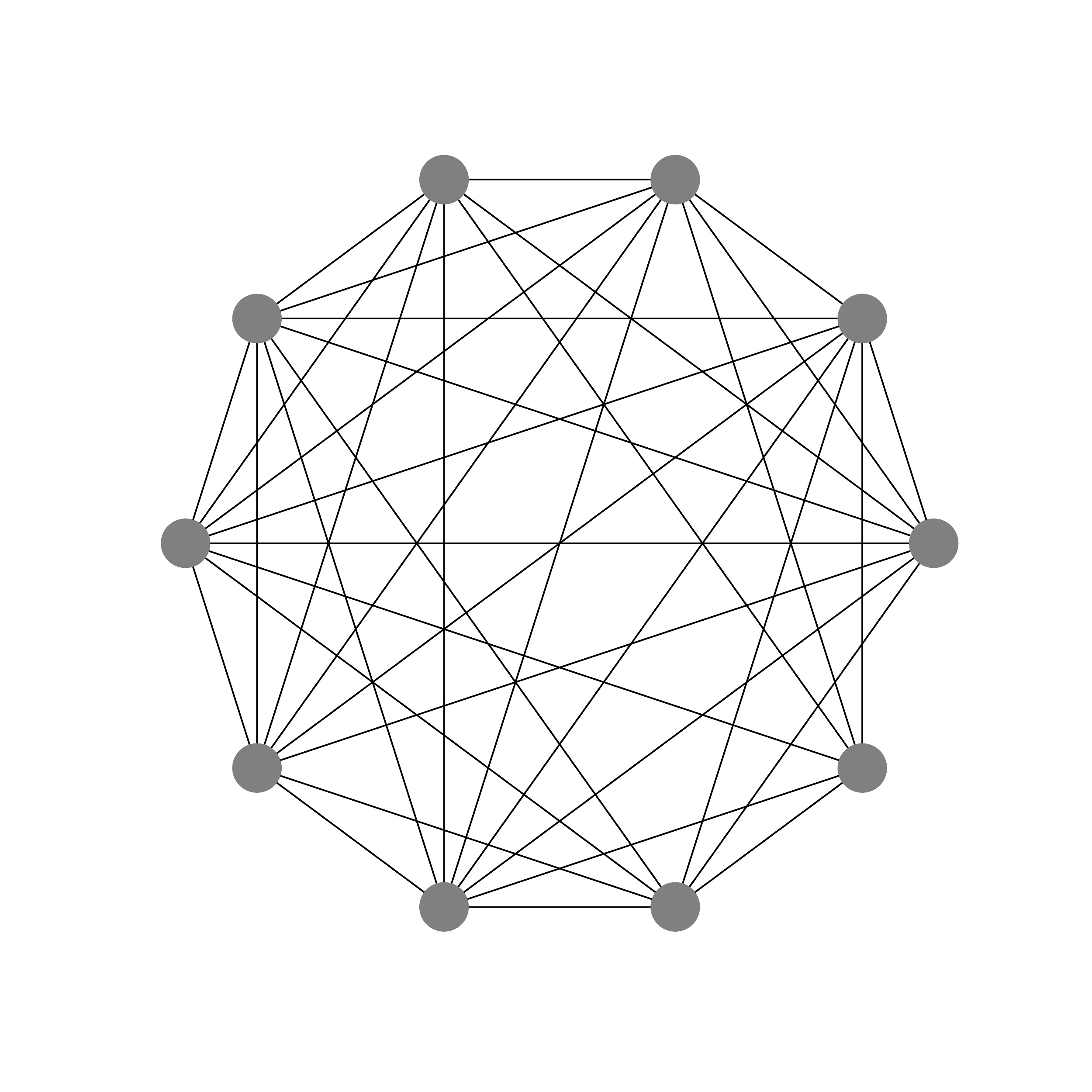}
		\end{minipage}
		\begin{minipage}{0.24\textwidth}
			\centering
			\caption[Examplary ER graphs with 10 vertices and distinct densities.]{Exemplary ER graphs with 10 vertices and edge probabilities $p=0.1$ (sparse graphs), $p=0.5$ (medium dense graphs), and $p=0.9$ (dense graphs).}
			\label{fig:expgraphs}
		\end{minipage}
	}
\end{figure*}

Based on this above results, we are now able to prove the main theorem by following the proof outline given in \Cref{sec:skeleton}.

\textit{Proof of Theorem~2.3.}
On an abstract level, the given neural network defines a class of functions $\Ff=\set{f_\theta:\Xx\to [0,1]:\theta \in \Theta}$ where the respective parameters $\theta\in\Theta$ are arranged in a specific manner. We leverage this function class and define an auxiliary binary class
$$
\Hh=\set{\indicator{\geq 0}(f(x)+wy+b):f\in\Ff, w,b\in\R)} ,
$$
which contains function on $\Xx\times \R$ and introduces a new weight and bias $w,b\in\R$. From \Cref{lem:numberparamssinglehiddenlayer}, we know that $\Hh$ has $W+2=a^2+3a+3$ adjustable parameters.

The function class $\Hh$ defines a concept class $\set{C_h\subseteq \Xx\times \R:h\in\Hh}$. In order to derive the computational complexity of deciding whether an input belongs to a given concept, we rely on ``the worst case'', \ie an input of maximal length $(x_1,\ldots,x_b,y)\in\Xx\times\R$. Then, in each time step, the RNN performs $2a^2-a$ arithmetic operations to compute $W^Th^{i-1}$, an additional $a$ arithmetic operations to compute $U^Tx_i$, and $4a$ operations add the components and apply the rectifier activation function. Hence, we require $b(2a^2+4a)$ operations to process the whole input $x=(x_1,\ldots,x_b)$ to a final hidden state $h^b$.
The original output unit of the real-valued ReLU RNN then performs an additional $2a$ operations to compute the final value $f(x)$, which functions as an input to the new thresholded output unit. This output unit performs additional $5$ operations in order to compute $h(x,y)$ and thereby decides whether $(x_1,\ldots,x_b,y)\in C_h$ as the result is already thresholded. In total, the computational complexity therefore equals
$$
T=T(W,b)=b(2a^2+4a)+2a+5.
$$

By application of \Cref{thm:vccomputationalcomplexity}, we can follow that
\begin{align*}
&\text{VCdim}(\Hh)\leq 2(W+2)(2T(W,b)+\log_2(8e))\\
&=2(a^2+3a+3)(2b(2a^2+4a)+4a+10+\log_2(8e) ,
\end{align*}
which, by further application of Lemma~\ref{lem:VCpseudo} and Theorem~2 in the main paper, implies the desired sample complexity bound.

\subsection{Proof of Theorem~2.4}

We extend the proof of Theorem~2.3 in order to cover multi-layer ReLU RNN and, therefore, assume $d$ recurrent layers of widths $(a_1,\ldots,a_d)$. The following lemma presents the number of adjustable parameters in this case.

\begin{lemma}
	\label{lem:multiW}
	The described RNN has $$W=\sum_{i=1}^{d}(a_i^2+2a_i)+a_d+1$$ trainable weights and biases.
	\begin{proof}
		We start by considering the first hidden layer. Like in the single hidden layer case, the RNN relies on $a_1^2+2a_1$ weights and biases in order to compute the output of the layer. Then, the whole $a_1$-dimensional hidden state is treated as an input sequence to the second hidden layer which incorporates the sequence one by one into a new $a_2$-dimensional hidden state. As before, this requires $a_2^2+2a_2$ new weights and biases. By the same procedure, the initial input traverses through the network until the last recurrent layer reaches the final hidden state $h_d$. Up to this point a total of $\sum_{i=1}^{d}a_1^2+2a_i$ adjustable parameters is used, and additional $a_d+1$ parameters are comprised in the linear output unit.
	\end{proof}
\end{lemma}

\textit{Proof of Theorem~2.4.}
Let $\Ff=\set{f_\theta:\Xx\to[0,1]:\theta\in\Theta}$ be the function class defined by the ReLU RNN. We define the auxiliary function class $\Hh$ on $\Xx\times \R$ as before and observe that $\Hh$ has $W+2=\sum_{i=1}^{d}(a_i^2+2a_i)+a_d+3$ adjustable weights and biases by \Cref{lem:multiW}.

Given a maximal length input $(x_1,\ldots,x_b,y)\in\Xx\times\R$, we again require $b(2a_1^2+4a)$ operations to process $x=(x_1,\ldots,x_b)$ to a final hidden state of the first recurrent layer $\textbf{h}_{\textbf{1}}=\textbf{h}_{\textbf{1}}^b$. In each subsequent recurrent layer $j=2,\ldots,d$, the network performs $2a_j-a_j$ arithmetic operations to compute $W_\textbf{j}^T\textbf{h}_{\textbf{j}}^i$, an additional $a_j$ operations in order to compute $U_{\textbf{j}}^T\textbf{h}_{\textbf{j-1}i}$, and $4a_j$ operations for addition and activation function. Here, $i$ indicates a single time-steps of the process. Whereas the first layer had an input sequence of length $b$, all other recurrent layers operate on the hidden state of the previous layer and iterate thus through $a_{j-1}$ time steps. This implies that by the time the last hidden state of the last recurrent layer is generated, the neural network performed a total of
$$
b(2a_1^2+4a_1)+\sum_{j=1}^{d-1}a_j(2a_{j+1}^2+4a_{j+1})
$$
operations. Additional $2a_d$ operations are required in order to output $f(x)$, and $5$ more operations are performed by the new thresholded output unit. The claim now follows directly by Lemma~\ref{lem:VCpseudo}, \Cref{thm:vccomputationalcomplexity} and Theorem~2 in the main paper.

%%%%%%%%%%%%%%%%%%%%%%%%%%%%%%%%%%%%%%%%%%%%%%%%%%%%%%%%%%%%%%%%%%%%

\section{Experimental details}
\subsection{Data generation}

We rely upon Erd\"os-Renyi (ER) random graphs for both training and testing. ER graphs are defined by a given number of vertices $n$ and a constant edge probability $p$ with which each two vertices independently form an edge. This procedure yields a probability distribution on the set of graphs with $n$ vertices \cite{diestel_graph_2000}. Thereby, each graph with the same number of edges $m$ is assigned the same probability of
\begin{equation*}
p^m(1-p)^{\binom{n}{2}-m},
\end{equation*}
such that the expected number of edges is $p\binom{n}{2}$.

We experiment with ER graphs that differ in their density, \ie, their expected number of edges relative to the possible number of edges. Examples are depicted in \Cref{fig:expgraphs}. We create three datasets containing graphs with different densities. First, we set the edge probability to $p=0.1$ and, thereby, yield particularly sparse graphs. Second, we set $p=0.5$, which evidently assigns every graph with $n$ vertices the same probability, and, third, we set $p=0.9$ to obtain dense graphs. Each of the three datasets contains 150,000 graphs and, finally, we merge the three datasets yielding a set of graphs with a mixture of sparse and dense graphs.

Data generation is parallelized into 48 processes. We limit our analysis to comparatively small graphs and different sizes at different density levels due to computational limitations, since it involves solving the NP-hard ECCN. Exact ECCNs are determined with a search algorithm that (1)~determines all maximal cliques with the Bron-Kerbosch algorithm \cite{bron_algorithm_1973}, and (2)~tries to cover the graphs' edges with an increasing number of maximal cliques in a brute-force manner. %\footnote{Our implementation of this ECCN algorithm is part of the supplementary material.} 
This is an inherent challenge when comparing RNN predictions to the exact ECCN and, because of this reason, labeling instances larger than 18 vertices becomes computationally intractable

\subsection{Training details}

Each of the proposed neural networks is analyzed separately on three distinct datasets. Therefore, we divide each data set into a training dataset (70\,\%), a validation dataset (10\,\%), and a test dataset (20\,\%). Both training and validation data are used for selecting the network parameters, while the test data is reserved for evaluating the final network. We ensured that the graphs of different sizes are distributed uniformly in the datasets. In addition, we pad the vector representations of graphs with zeros such that all instances in a dataset have the same length.  

In all experiments, neural networks are trained on the respective training data set by a mini-batch-based variant of the stochastic gradient descent algorithm that uses the adaptive learning rate algorithm \emph{Adam} \cite{kingma_adam:_2014}. For the feed-forward network, we implemented a three-layered architecture and tried different learning rates of 0.01, 0.001 and 0.0001. For the RNN and LSTM network, we varied the size of the hidden state across 256, 512 and 1024.

The training data is used for fitting of the network, whereas the validation data is solely used to choose a network specification that successfully generalizes to unseen samples. This ensures that we avoid overfitting, while still achieving a high approximation quality of our results. For this purpose, we further draw upon an early stopping technique that monitors the loss on the validation data set with a maximum of 2000 epochs and patience of 20.

\end{document}